\documentclass[conference]{IEEEtran}
\IEEEoverridecommandlockouts
\usepackage{cite}
\usepackage{amsmath,amssymb,amsfonts}
\usepackage{algorithmic}
\usepackage{graphicx}
\usepackage{textcomp}
\usepackage{xcolor}
\usepackage{multirow}
\usepackage{arydshln}
\usepackage{todonotes}
\usepackage[backref, colorlinks,urlcolor=blue]{hyperref}

\def\BibTeX{{\rm B\kern-.05em{\sc i\kern-.025em b}\kern-.08em
    T\kern-.1667em\lower.7ex\hbox{E}\kern-.125emX}}
    
\usepackage{url}
\newcommand{\etal}{\emph{et al. }}

\begin{document}
    
\title {Evaluation of a Dual Convolutional Neural Network Architecture for Object-wise Anomaly Detection in Cluttered X-ray Security Imagery}

\author{\IEEEauthorblockN{Yona Falinie A. Gaus$^1$, Neelanjan Bhowmik$^1$, Samet Ak\c{c}ay$^1$, \\ Paolo M.  Guill\'en-Garcia$^{1,2}$, Jack W. Barker$^1$, Toby P. Breckon$^1$}
\IEEEauthorblockA{$^1$\textit{Dept. of Computer Science}, \textit{Durham University}, Durham, UK \\
$^2$\textit{Dept. of Biomedical Eng.}, \textit{Universidad Polit\'ecnica de Chiapas}, Chiapas, Mexico}

}

\maketitle

\begin{abstract}
    X-ray baggage security screening is widely used to maintain aviation and transport security. Of particular interest is the focus on automated security X-ray analysis for particular classes of object such as electronics, electrical items and liquids. However, manual inspection of such items is challenging when dealing with potentially anomalous items. Here we present a dual convolutional neural network (CNN) architecture for automatic anomaly detection within complex security X-ray imagery. We leverage recent advances in region-based (R-CNN), mask-based CNN (Mask R-CNN) and detection architectures such as RetinaNet to provide object localisation variants for specific object classes of interest. Subsequently, leveraging a range of established CNN object and fine-grained category classification approaches we formulate within object anomaly detection as a two-class problem (anomalous or benign). While the best performing object localisation method is able to perform with  97.9\% mean average precision (mAP) over a six-class X-ray object detection problem, subsequent two-class anomaly/benign classification is able to achieve 66\% performance for within object anomaly detection. Overall, this performance illustrates both the challenge and promise of object-wise anomaly detection within the context of cluttered X-ray security imagery. 
\end{abstract}

    %%%%%%%%%%%%%%%%%%%%%%%%%%%%%%%%%%%%%%%%%%%%%%%%%%%%%%%%%%%%%%%%

\section{Introduction}

X-ray baggage security screening is widely used to maintain aviation and transport security, itself posing a significant image-based screening task for human operators reviewing compact, cluttered and highly varying baggage contents within limited time-scales. With both increased passenger throughput in the global travel network and an increasing focus on broader aspects of extended border security (e.g. freight, shipping postal), this poses both a challenging and timely automated image classification task.

%%%%%%%%%%%%%%%%%%%%%%%%%%%%%%%%%%%%%%%%%%%%%%%%%%%%%%%%%%%%%%%%

To facilitate effective screening, threat detection via scanned X-ray imagery is increasingly employed to provide a non-intrusive, internal view of scanned baggage, freight, and postal items. This produces colour-mapped X-ray images which correspond to the material properties detected via the dual-energy X-ray scanning process \cite{Rogers2016ReviewXray}. While current automatic threat detection within X-ray security screening concentrates on material discrimination for explosive-related threats \cite{Rogers2016ReviewXray}, a growing body of work illustrates the potential of CNN architectures for broader object based threat detection \cite{Akcay2016Xray, Akcay2017Xray, Akcay2018Xray}. In both occurrences, threat detection performance must be characterised by high detection and low false alarm rates for operational viability. 

Within this context, of particular interest are electronics, electrical items and liquids \cite{government2015digital}. Not only do these items come in many evolving variants but they are additionally packed in complex and cluttered surroundings leading to a complex X-ray image interpretation problem.

\begin{figure}[t]
    \includegraphics[width=\linewidth]{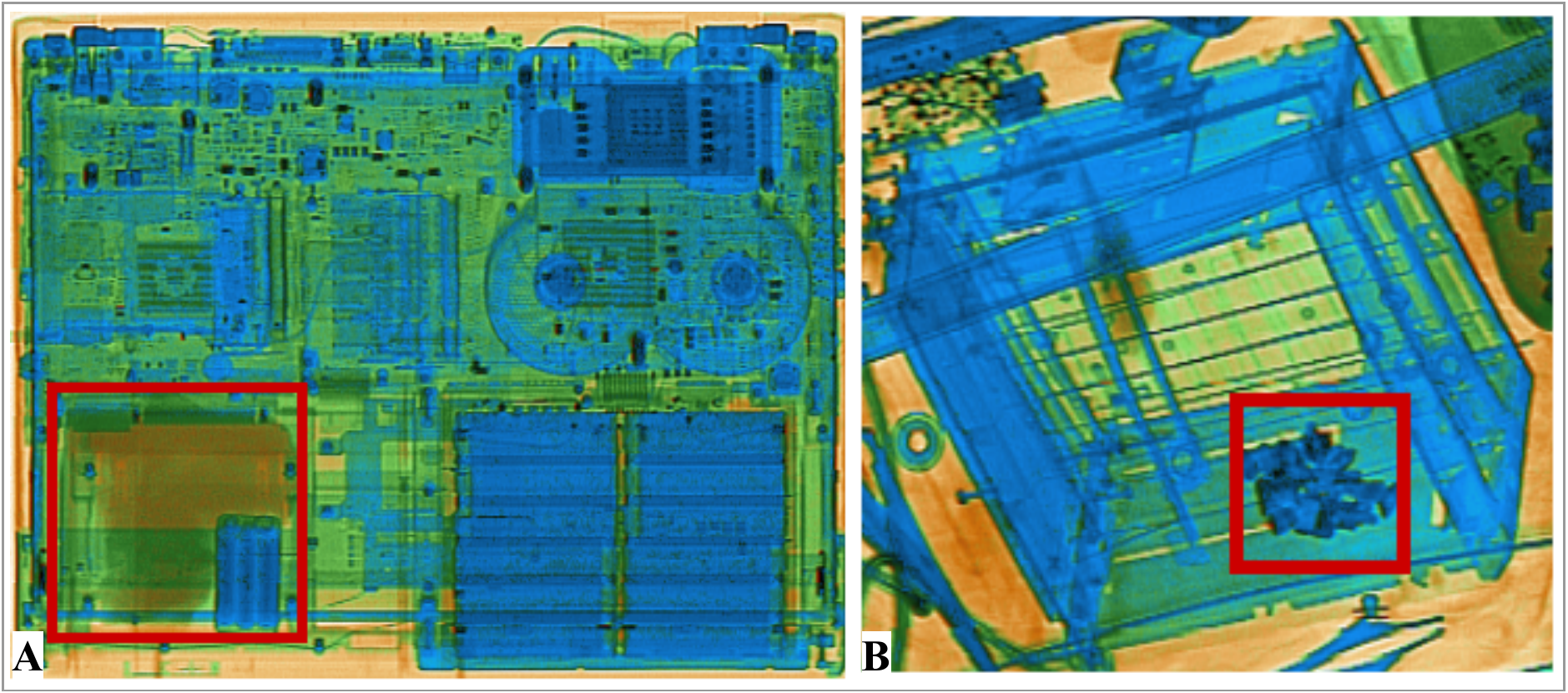}
    \caption{X-ray security imagery of exemplar electronics items
    with a highlighted (red box) concealed anomalous region in (A) laptop and (B) toaster.}
    \label{Fig:xray_inert_laptop_example}
\end{figure} 

Whilst existing security scanners use dual-energy X-ray for materials discrimination, and highlight specific image regions matching existing threat material profiles \cite{singh2003explosives, wells2012review}, the detection of generalized anomalies within complex items remains challenging \cite{korupski2018evaluation} (e.g. Fig. \ref{Fig:xray_inert_laptop_example}).

Within machine learning, anomaly detection involves learning a pattern or distribution of normality for a given data source and thus detecting significant deviations from this norm \cite{Patcha2007:AnomalyOverview}. Anomaly detection is an area of significant interest within computer vision, spanning biomedical imaging \cite{Schlegl2017:AnomalyGAN} to video surveillance \cite{Kiran2018:AnomalyVideo}. In our consideration of X-ray security imagery, we are looking for abnormalities that indicate concealment or subterfuge whilst working against a real-world adversary who may evolve their strategy to avoid detection. Such anomalies may present (or conceal) themselves within appearance space in the form of an unusual shape, texture or material density (i.e. dual-energy X-ray colour) \cite{greenemeier_2010}. Alternatively, they may present themselves in a semantic form, where the appearance of unfamiliar objects either globally or locally within the X-ray image \cite{brown_2018}.

Considering the notable challenge of detecting such subtle anomalies globally within the image, we instead follow a human-like approach to illustrate an automated pipeline for the object-wise screening of such items - \textit{focus on locating the object within the scene (image) first, then determine if the object is anomalous or not ?} 

By leveraging recent advances in object detection and classification in X-ray security imagery \cite{Akcay2016Xray, Akcay2017Xray, Akcay2018Xray}, we propose a dual CNN architecture to firstly isolate liquid and electrical objects by type and subsequently screen them for abnormalities. The main contribution of this work is a dual CNN architecture for object-wise anomaly detection, which jointly leverages state-of-the-art joint object detection and segmentation \cite{he2017maskrcnn} for first stage object localisation and subsequently considers second stage anomaly detection as a simple two-class CNN classification problem within cluttered X-ray security imagery.

    %%%%%%%%%%%%%%%%%%%%%%%%%%%%%%%%%%%%%%%%%%%%%%%%%%%%%%%%%%%%%%%%

\section{Related Work}
\label{Section:LR}
There has been a steady increase in research work considering object based threat detection in X-ray baggage security imagery.
Rogers \etal \cite{Rogers2016ReviewXray} performs a comprehensive review of the field, including baggage and cargo imagery. 
In this section, we will focus on supervised learning for automated threat detection and anomaly detection within X-ray security imagery.

%%%%%%%%%%%%%%%%%%%%%%%%%%%%%%%%%%%%%%%%%%%%%%%%%%%%%%%%%%%%%%%%

\subsection{Automated Threat Detection in X-ray Imagery}
\label{SubSection:Automatic threat detection in X-ray imagery}
Early work on X-ray security images is based on hand-crafted features such as Bag-of-Visual-Words (BoVW), which is applied together with a classifier such as a Support Vector Machine (SVM), achieving a performance of 0.7 recall, 0.29 precision, and 0.57 average precision \cite{bastan2011xray}. Turcsany \etal \cite{turcsany2013xray} extend the approach by using BoVW with SURF descriptor and SVM classifier yields 0.99 true positive and 0.04 false positive rates. 
Subsequently, BoVW is further evaluated for the single and dual-view X-ray images \cite{Bastan2013ObjectRI}, with optimal average precision achieved for firearms (0.95) and laptops (0.98). The various feature point descriptors within BOVW is explored thoroughly in the work of  \cite{kundegorski2016xray}, where the best the combination achieves 94.0\% accuracy with two classes of firearm detection using an SVM classifier. 

Recent CNN-based deep learning architectures \cite{Krizhevsky2017ImageNet,he2016residual,szegedy2016inception,howard2017mobilenets} have significantly improved the object detection in X-ray security imaging \cite{Akcay2018Xray, Jaccard2016a, Mery2017Xray}. Earlier work on CNN in X-ray imaging \cite{Akcay2016Xray} explore the use of transfer learning from another network trained on a classification task. Experiments show that CNN with transfer learning achieves superior performance, 99.26\% on true positive and 4.08\% on false positive, only by fine-tuning the network. Broader experimentation in \cite{Akcay2018Xray} empirically proves the superiority of fine-tuned CNNs over the classical machine learning algorithms.

%%%%%%%%%%%%%%%%%%%%%%%%%%%%%%%%%%%%%%%%%%%%%%%%%%%%%%%%%%%%%%%%
\begin{figure*}[t]
    \centerline{\includegraphics[width=\linewidth]{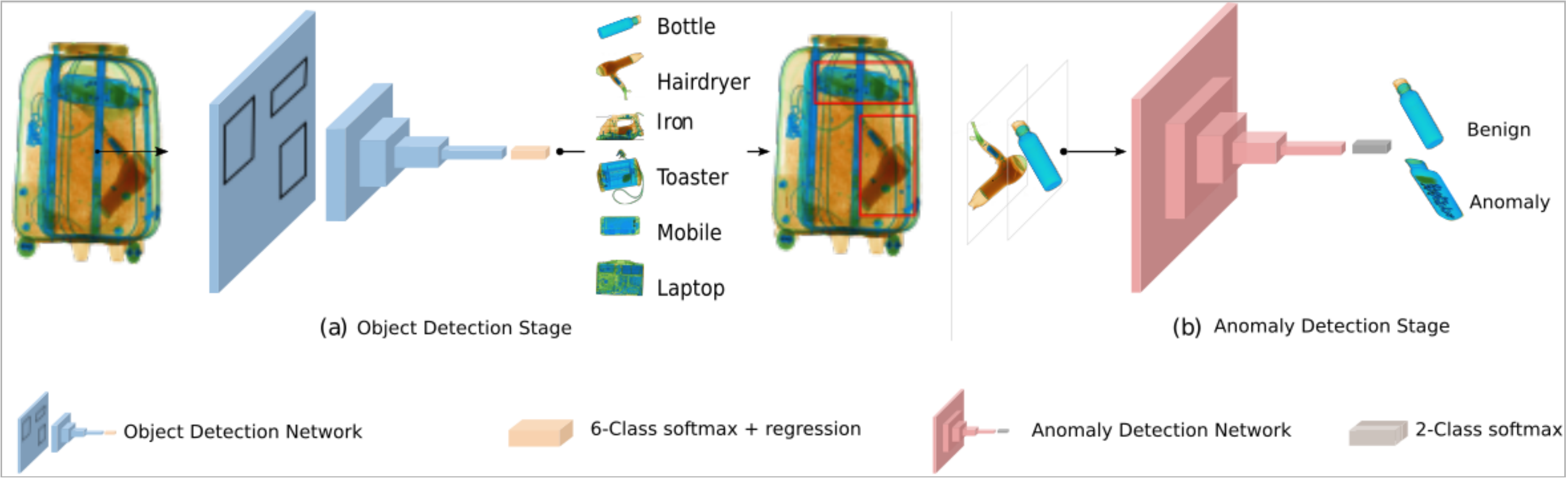}}
    \caption{Our dual CNN architecture for object-wise anomaly detection in complex X-ray security imagery.}
    \label{Figure:2-stage arch}
\end{figure*}
%%%%%%%%%%%%%%%%%%%%%%%%%%%%%%%%%%%%%%%%%%%%%%%%%%%%%%%%%%%%%%%%

\subsection{Automated Anomaly Detection in X-ray Imagery}
\label{SubSection:Automatic anomaly detection in X-ray imagery}

Sterchi \etal \cite{Sterchi2017} show that security officers are able to detect the abnormality better when they focus on detecting each object in the bag as benign rather than concentrating on threat items. By the same analogy, the anomaly detection algorithms proposed in the field are trained on benign samples to learn what is normal and tested on both normal and abnormal images to detect threats. 

Prior work on appearance and semantic anomaly detection, has considered unique feature representation as a critical component for detection within cluttered X-ray imagery \cite{griffin2018unexpected}. Early work on anomaly detection in X-ray security imagery \cite{zheng2013xray}, implements block-wise correlation analysis between two temporally aligned scanned X-ray images. More recently \cite{Andrews2016:AnomalyAutoEncoder}, anomalous X-ray items within freight containers have been detected using auto-encoder networks, and additionally via the use convolutional neural network (CNN) extracted features as a learned representation of normality across stream-of-commerce parcel X-ray images \cite{griffin2018unexpected}. Andrews \etal \cite{andrews2017xray} propose representational-learning for anomaly detection within cargo container imagery. In a similar vein, the work of \cite{Akcay2018GANomaly} focuses on the use of a novel adversarial training architecture to detect anomalies based on high reconstruction errors produced by a generator network adversarially trained on non-anomalous (benign) stream-of-commerce X-ray imagery only. In follow-up work, \cite{Akcay2019} proposes another unsupervised anomaly detection approach, whereby the use of skip connected layer design allows to train much higher resolution images and optimising latent space within the discriminator network leads to significantly better results.

By contrast, here we consider a two-stage approach that first isolates potential objects of interest within the X-ray security image, as an object detection and classification problem (Section \ref{Subsection:Detection Strategies}), prior to secondary anomaly detection via application of CNN based image classification (Section \ref{Subsection:Classification Strategies}).

%%%%%%%%%%%%%%%%%%%%%%%%%%%%%%%%%%%%%%%%%%%%%%%%%%%%%%%%%%%%%%%%
    %%%%%%%%%%%%%%%%%%%%%%%%%%%%%%%%%%%%%%%%%%%%%%%%%%%%%%%%%%%%%%%%

\section{Proposed Approach}
\label{Sec:proposed approach}
Our dual CNN architecture performs two stages of analysis:- (a) primary object detection within the X-ray image (Section \ref{Subsection:Detection Strategies}); and (b) secondary classification of each detected object via a two-class, $\{anomaly, benign\}$, classification formulation (Section \ref{Subsection:Classification Strategies}). An overview of our overall dual CNN architecture is shown in Fig. \ref{Figure:2-stage arch}.

%%%%%%%%%%%%%%%%%%%%%%%%%%%%%%%%%%%%%%%%%%%%%%%%%%%%%%%%%%%%%%%%

\subsection{Detection Strategy}
\label{Subsection:Detection Strategies}
\begin{figure}[t]
 \centerline{\includegraphics[width=8.5cm]{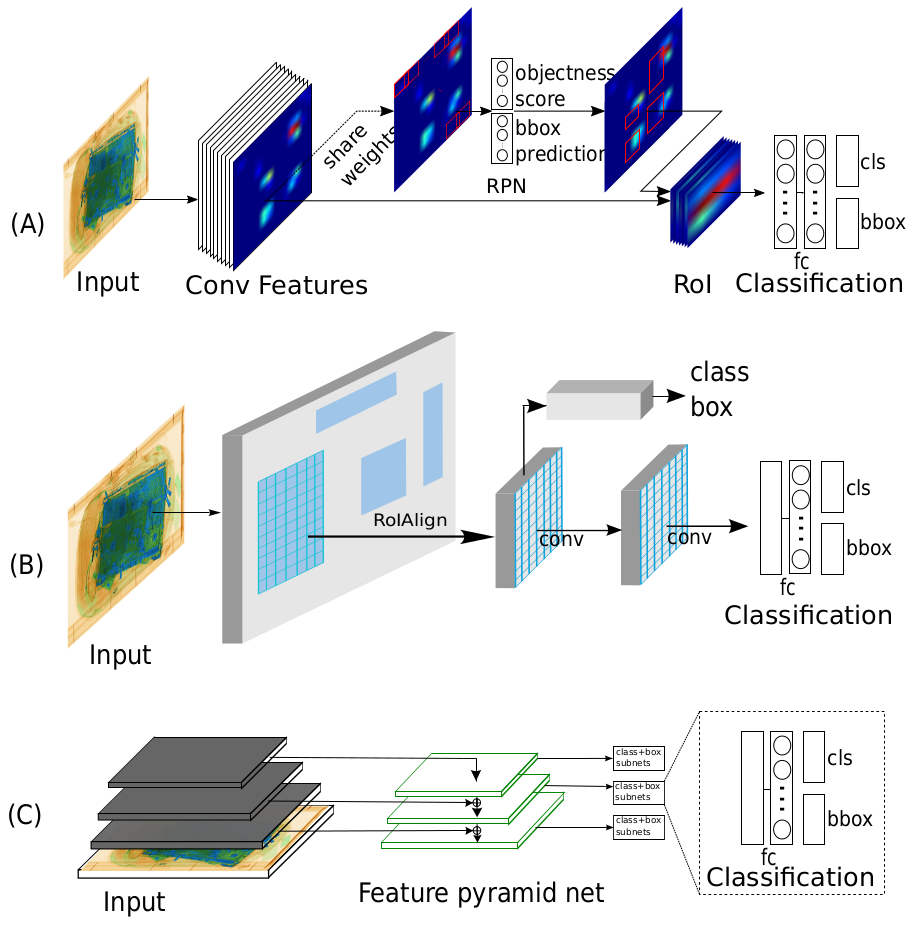}}
 \caption{Architecture of the CNN based detection approaches evaluated:
(A) Faster R-CNN \cite{ren2015fasterrcnn} (B) Mask R-CNN \cite{he2017maskrcnn} (C) RetinaNet \cite{lin2017focalloss}.}
 \label{Fig:cnnModels}
 \end{figure}

We consider a number of contemporary CNN frameworks for our primary object detection strategy to explore their applicability and performance for generalised object detection within the context of X-ray security imagery. Namely we consider {\it Faster R-CNN} \cite{ren2015fasterrcnn}, {\it Mask R-CNN} \cite{he2017maskrcnn} and {\it RetinaNet} \cite{lin2017focalloss} with internal architectures as illustrated in the Fig. \ref{Fig:cnnModels}. These are evaluated over a six class object detection and localisation problem comprising of \{{\it bottle, hairdryer, iron, toaster, mobile, laptop}\} items packed within cluttered X-ray security baggage imagery. 

{\it Faster R-CNN} is based on a two-stage internal architecture \cite{ren2015fasterrcnn}, as shown in Figure \ref{Fig:cnnModels}(A). The first stage consists of a Region Proposal Network (RPN) that proposes regions of interest to a secondary classification stage. The RPN consists of convolutional layers that generate set of anchors with different scales and aspect ratios, and predict their bounding box coordinates together with a probability score denoting whether the region is an object of interest or background. Anchors are generated by using a fixed set of nine standard axis-aligned bounding boxes in three different aspect ratios and three scales,
which are defined at every location of the feature maps. These features are then fed into objectness classification and bounding box regression layers.  Within the second stage, the objectness classification layer classifies whether a given region proposal is an object or a background region while a bounding box regression layer predicts object localisation, at the end of the overall detection process.

{\it Mask R-CNN} is an extension of the Faster R-CNN architecture for combined object localisation and instance segmentation of image objects \cite{he2017maskrcnn}. Mask-RCNN similarly relies on a region proposals which are generated via a region proposal network. Mask-RCNN follows the Faster-RCNN model of a feature extractor followed by this region proposal network, followed by an operation known as ROI-Pooling to produce standard-sized outputs suitable for input into a secondary classifier. The main differences between Mask-RCNN and Faster-RCNN rely on three factors. Firstly, Mask-RCNN replaces the ROI-Pooling operation used in Faster-RCNN with an operation called ROI-Align that allows very accurate instance segmentation masks to be constructed. Secondly, Mask-RCNN adds a network head (a small fully convolutional neural network) to produce the desired instance segmentation, as in Fig. \ref{Fig:cnnModels}(B). Finally, segmentation and classification label predictions are decoupled; the mask network head predicts the instance segmentation independently from the network head predicting the classification label for the object that is being segmented.

{\it RetinaNet} is a one-stage object detector proposed by Lin et al. \cite{lin2017focalloss}, where the author identified that class imbalance are the critical reasons why the performance of single stage detector architectures such as YOLO \cite{redmon2016you} and SSD \cite{liu2015ssd} lag behind two-stage detector architectures such as Faster R-CNN and Mask R-CNN. To improve the performance, RetinaNet employs a novel loss function called Focal Loss, which allows it to focus more on class imbalance samples. Using a one-stage network architecture with Focal Loss, RetinaNet achieves state-of-the-art performance in terms of accuracy and running time. Figure \ref{Fig:cnnModels}(C) depicts the overall architecture of RetinaNet, which is composed of a backbone network and two sub-networks. The backbone network is responsible for computing a convolutional feature map using the Feature Pyramid Network (FPN) over an entire input image. Subsequently, the first subnet performs label classification on the backbones output, while the second subnet performs convolution bounding box regression (i.e. localisation). The focal loss is applied as the loss function as shown in Fig. \ref{Fig:cnnModels}(C).

%%%%%%%%%%%%%%%%%%%%%%%%%%%%%%%%%%%%%%%%%%%%%%%%%%%%%%%%%%%%%%%%

\subsection{Classification Strategy}
\label{Subsection:Classification Strategies}

After detecting the candidate objects of interest within X-ray security imagery based on our detection strategy (Section \ref{Subsection:Detection Strategies}), our secondary classification strategy determines whether the object localised within the image is \{{\it anomaly, benign}\} as a two-class classification problem.

In doing so, we leverage transfer learning both from a set of seminal CNN object classification architectures (SqueezeNet \cite{Iandola2016SqueezeNet}, VGG-16 \cite{simonyan2014verydeep}, ResNet \cite{he2016residual}) pre-trained on ImageNet  \cite{Russakovsky2015ImageNet} following the X-ray classification methodology of \cite{Akcay2016Xray}, and a CNN architecture specifically designed for fine-grained (i.e. sub-class) classification tasks \cite{Wang2018LearningAD}.

We specifically adopt fine-grained classification for anomaly detection, where we define benign and anomalous as sub-categories (sub-classes) of the primary object type detected (Section \ref{Subsection:Detection Strategies}). Within the literature, fine-grained classification usually aims to distinguish subordinate visual categories to the main object class such as determining natural categories such as species of birds \cite{VanHorn2015iIrds} \cite{Wah2011Birds}, dogs \cite{Khosla2011finegrained} and plants \cite{Wegner2016Catalogue}. 

In the case of our \{{\it anomaly, benign}\} classification problem, the key to successful fine-grained classification lies in developing an automated method to accurately identify informative regions in an anomalous item, and whether each such region belongs to an anomalous region or benign region of the overall object. 

However, labelling the discriminate regions requires significant manual annotation and is therefore difficult to scale effectively. To avoid this issue, we specifically utilise fine-grained classification learning a discriminative filter bank within a CNN framework in an end-to-end manner without the need for explicit additional object annotation \cite{Wang2018LearningAD}. This approach enhances mid-level representational learning within the CNN architecture, by learning a set of convolution filters such that each is initialised and discriminatively trained in order to capture highly discriminative sub-image patches. Based on the 
the VGG-16 network architecture \cite{simonyan2014verydeep}, filters are additionally added at the $10^{th}$ convolutional layer representing image patches as small as $92\times92$ with a stride of 8 \cite{Wang2018LearningAD}.

%%%%%%%%%%%%%%%%%%%%%%%%%%%%%%%%%%%%%%%%%%%%%%%%%%%%%%%%%%%%%%%%
    %%%%%%%%%%%%%%%%%%%%%%%%%%%%%%%%%%%%%%%%%%%%%%%%%%%%%%%%%%%%%%%%
\begin{table*}[t]
  \centering
  \caption{object detection results for faster r-cnn, mask r-cnn and retinanet for dual cnn architecture. class names indicates corresponding average precision (AP) of each class, and mAP indicates mean average precision of the classes.}
  \begin{tabular}{llllllllc}
  \hline
  \multirow{2}{*}{Model} &  \multirow{2}{*}{\shortstack[l]{Network \\ configuration}} & \multicolumn{6}{c}{Average precision} & \multirow{2}{*}{mAP} \\ \cline{3-8}
  & & Bottle & Hairdryer & Iron & Toaster & Mobile & Laptop &   \\  \hline \hline
  
  \multirow{2}{*}{Faster R-CNN \cite{ren2015fasterrcnn}} & ResNet$_{101}$  &96.7&97.5 &98.0 &98.2 &96.4 &97.3 &97.4 \\ 
                
    & ResNet$_{50}$       &95.5    &96.4        & 97.6          &94.0     & 94.3        &   96.8     &   95.8                 \\\hline
  
  \multirow{2}{*}{Mask R-CNN\cite{he2017maskrcnn}}   & ResNet$_{101}$      & \textbf{99.4}   &92.2        &\textbf{100.0}           & \textbf{100.0}     & 96.5        & \textbf{99.6}       &  \textbf{97.9}                  \\ 
    & ResNet$_{50}$       &97.8    & 90.8       & 99.9          &99.6     & 95.5        &  98.3      &    96.9                \\ \hline
  
  \multirow{2}{*}{RetinaNet\cite{lin2017focalloss}} & ResNet$_{101}$ & 95.2 & \textbf{99.2} & 98.6         & 98.5     &   \textbf{97.7}      &86.7        &   95.9                 \\ 
  
      &ResNet$_{50}$       &98.3     & 98.6       &98.9           & 96.7     & 87.5        &  88.5     &  94.8                  \\ \hline
  \end{tabular}
  \label{Table:detection AP and mAP}
  \end{table*}
  \begin{figure*}
      \includegraphics[width=\linewidth]{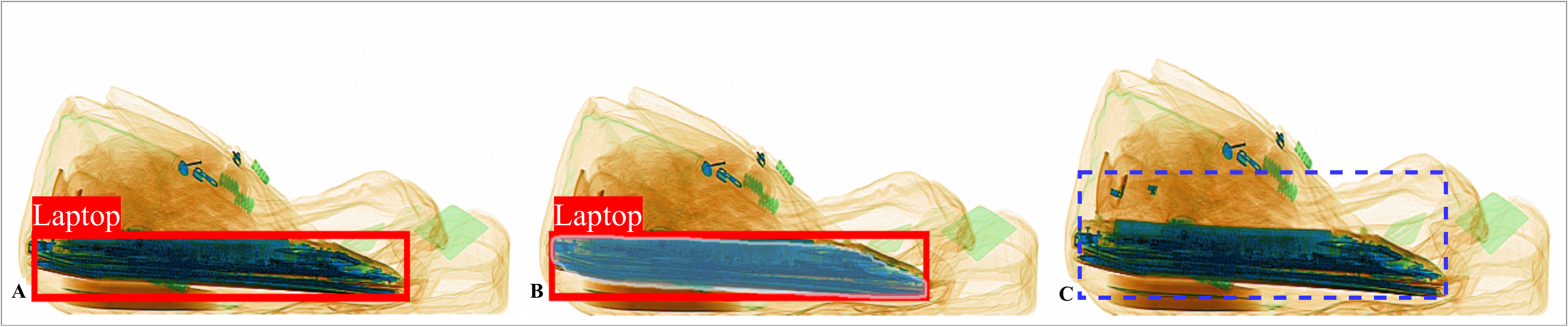}
      \caption{Exemplar image cases where RetinaNet (C) fails to detect an object (laptop in blue dashed box) in X-ray image, while Faster R-CNN (A) and Mask R-CNN (B) are able to detect the object.}
      \label{Fig:model_detection_compare}
  \end{figure*} 
  \begin{table*}[t]
  \centering
  \caption{anomaly classification via varying cnn architectures (squeezenet, vgg, and resnet) with and without the pre-localization offered by the proposed dual cnn architecture.}
  \begin{tabular}{lllllllll}
  \hline
  Object Detection & Model & Network configuration & A & P & R & F1 & TP(\%) & FP(\%)\\ \hline \hline
  \multirow{5}{*}{\begin{tabular}[c]{@{}l@{}}Dual CNN\\ (pre-localization)\end{tabular}}      & \multirow{4}{*}{\begin{tabular}[c]{@{}l@{}}Classification via CNN\end{tabular}}  & ResNet$_{18}$ & \textbf{0.66}  & 0.67  & 0.58  &0.30 & 58.11 &26.56 \\  
                         &                            & ResNet$_{50}$              &  0.66                        &   0.67                     &0.59   &0.63 & 59.25& 27.67  \\ 
                         
          &   & SqueezeNet &  0.59                         &       0.57                 & \textbf{0.77} &0.57 &\textbf{76.86} & 57.16  \\ 
              &   & VGG-16  & 0.59                         &   \textbf{0.74}                      &0.75  &0.56 &74.51 &55.31   \\ \cline{2-9}                    
   & Classification via Fine-Grained  & VGG-16 &0.64  &0.62   &0.70  &\textbf{0.66}  &70.00  &58.00 \\ \hline 
  
  \multirow{4}{*}{\begin{tabular}[c]{@{}l@{}}Full Image\\ (no localization)\end{tabular}}      & \multirow{4}{*}{\begin{tabular}[c]{@{}l@{}} Classification via CNN\end{tabular}}  &ResNet$_{18}$ & 0.57   & 0.57  & 0.58  &0.50&58.19  &43.42 \\ 
  
     &                            &ResNet$_{50}$ &     0.59                    &   0.58                     & 0.61  & 0.58& 61.24& 42.81  \\ 
     
        &                            & SqueezeNet          &   0.58                   &   0.72                     &0.27   &0.53 &26.76 &\textbf{10.36} \\ 
        &                            & VGG-16              &     0.52                     &0.53                        &0.23   &0.43 &22.86 & 19.08  \\ \hline
  
  \end{tabular}
  \label{Table:binary class}
  \end{table*}
  
  \section{Experimental Setup}
  \label{Section:Experimental Setup}
  In this section, we introduce the dataset, evaluation criteria and CNN training details used in this work.
  
  %%%%%%%%%%%%%%%%%%%%%%%%%%%%%%%%%%%%%%%%%%%%%%%%%%%%%%%%%%%%%%%%
  
  \subsection{Dataset}
  \label{Subsection:Datasets}
  We construct our dataset using single-view conventional X-ray imagery with associated false colour materials mapping from dual-energy \cite{Mouton2015ARO}. Our X-ray images consist of benign and anomalous items, such as a laptop, mobile, toaster, iron, hairdryer and bottle. To introduce anomalies, we insert marzipan, screws, metal plates, knives and alike inside these objects as depicted in Fig. \ref{Fig:xray_inert_laptop_example}. All X-ray imagery is gathered locally used a Gilardoni dual-energy X-ray scanner (FEP ME 640 AMX, \cite{gilardoni_scanner}).
  
  Each of the anomalous items is placed inside various cluttered baggage items, which cover the full range and dimensions found in aviation cabin baggage, ensuring the set of bags is a good representation of such items typically presented at the aviation checkpoint security. In total, the number of X-ray images after scanning of each bags is 3534 images.

  %%%%%%%%%%%%%%%%%%%%%%%%%%%%%%%%%%%%%%%%%%%%%%%%%%%%%%%%%%%%%%%%
  
  \subsection{Evaluation Criteria}
  \label{Subsection:Evaluation-Criteria}
  
  For object detection, the performance of the models is evaluated by mean average precision (mAP), as used in the seminal object detection benchmark work of \cite{lin2014coco}. In order to calculate mAP, we calculate the area of intersection over union for the given ground truth and detected bounding box for each detection as:
  
  \begin{equation}
  \label{Eq:area_of_intersection}
  \Psi(B_{gt_i}, B_{dt_i})=\frac{Area(B_{gt_i}\cap B_{dt_i})}{Area(B_{gt_i}\cup B_{dt_i})}
  \end{equation}
  
  where $B_{gt_i}$ and $B_{dt_i}$ are ground truth and detected bounding box for detection $i$, respectively. Assuming each detection as unique, and denoting the area as $\Psi(B_{gt_i}, B_{dt_i})$,
  we then threshold it by the range of  $\theta=.50:.05:.95$ giving the logical $b_i$, where:
  \begin{equation}
  \label{Eq:logical}
    b_i=\begin{cases}
      1, &  \theta_{min}<\Psi(B_{gt_i}, B_{dt_i})<\theta_{max} \\
      0, &  a_{i}<\theta_{min}.
    \end{cases}
  \end{equation}
  Given both true positive and false positive as $t_i$  and $f_i$, where:
  \begin{equation}
  \begin{aligned}
  \label{Eq:TP_FP}
  t_{i}= & t_{i-1}+ b_{i}  \\
  f_{i}= & t_{i-1}+(1-b_{i})
  \end{aligned}
  \end{equation}
  The precision $p_i$ and recall $r_i$ curves can be calculated as:
  \begin{equation}
  \begin{aligned}
  \label{Eq:Precision_Recall}
  p_{i}= & \frac{t_{i}}{t_{i}+f_{i}}  \\
  r_{i}= & \frac{t_{i}}{n_{p}}
  \end{aligned}
  \end{equation}
  where $n_p$ is the number of positive samples. We can calculate average precision (AP) based on the area under the curve of precision versus recall:
  \begin{equation}
  \label{Eq:AP}                                                                                  
  AP=\sum_{i}^{n_d}p_i\bigtriangleup r
  \end{equation}
  Subsequently, we can get the value of mAP by averaging AP values for all classes, $C$:
  \begin{equation}
  \label{Eq:mAP}                                                       mAP=\frac{1}{C}\sum_{c=1}^{C}AP_{c}
  \end{equation}
  
  For anomaly detection via classification, our model performances are evaluated in terms of Accuracy (A), Precision (P), Recall (R), F-score (F1\%), True Positive (TP\%), and False Positive (FP\%).
  
  %%%%%%%%%%%%%%%%%%%%%%%%%%%%%%%%%%%%%%%%%%%%%%%%%%%%%%%%%%%%%%%%
  
  \subsection{Training Details}
  \label{Subsection:training-details}
  
  In our experiments, we use the CNN implementation of \cite{Girshick2018Detectron} for our primary object detection approach. Our models are trained on a GTX 1080Ti GPU, optimised by Stochastic Gradient Descent (SGD) with a weight decay of 0.0001, the learning rate of 0.0 and termination at 180k iterations. 
  
  The ResNet$_{50}$ and  ResNet$_{101}$\cite{he2016residual} are used as a network backbone for detection. When using ResNet$_{101}$ as the backbone, we drop the learning rate by half and double the length of the training schedule to facilitate training within the memory footprint of the available GPU. We split the datasets into training (60\%), validation (20\%) and test sets (20\%) such that each split has similar class distribution. We also perform scaling and horizontal flipping to each sample to augment the datasets during training. 
  All experiments are initialised with ImageNet pre-trained weights for their respective model \cite{deng2009imagenet}.  For Faster R-CNN and Mask R-CNN, the batch size is set to 512 for the RPN. 
  
  %%%%%%%%%%%%%%%%%%%%%%%%%%%%%%%%%%%%%%%%%%%%%%%%%%%%%%%%%%%%%%%%

    %%%%%%%%%%%%%%%%%%%%%%%%%%%%%%%%%%%%%%%%%%%%%%%%%%%%%%%%%%%%%%%%
\begin{figure*}
    \centering
    \includegraphics[width=16.5cm]{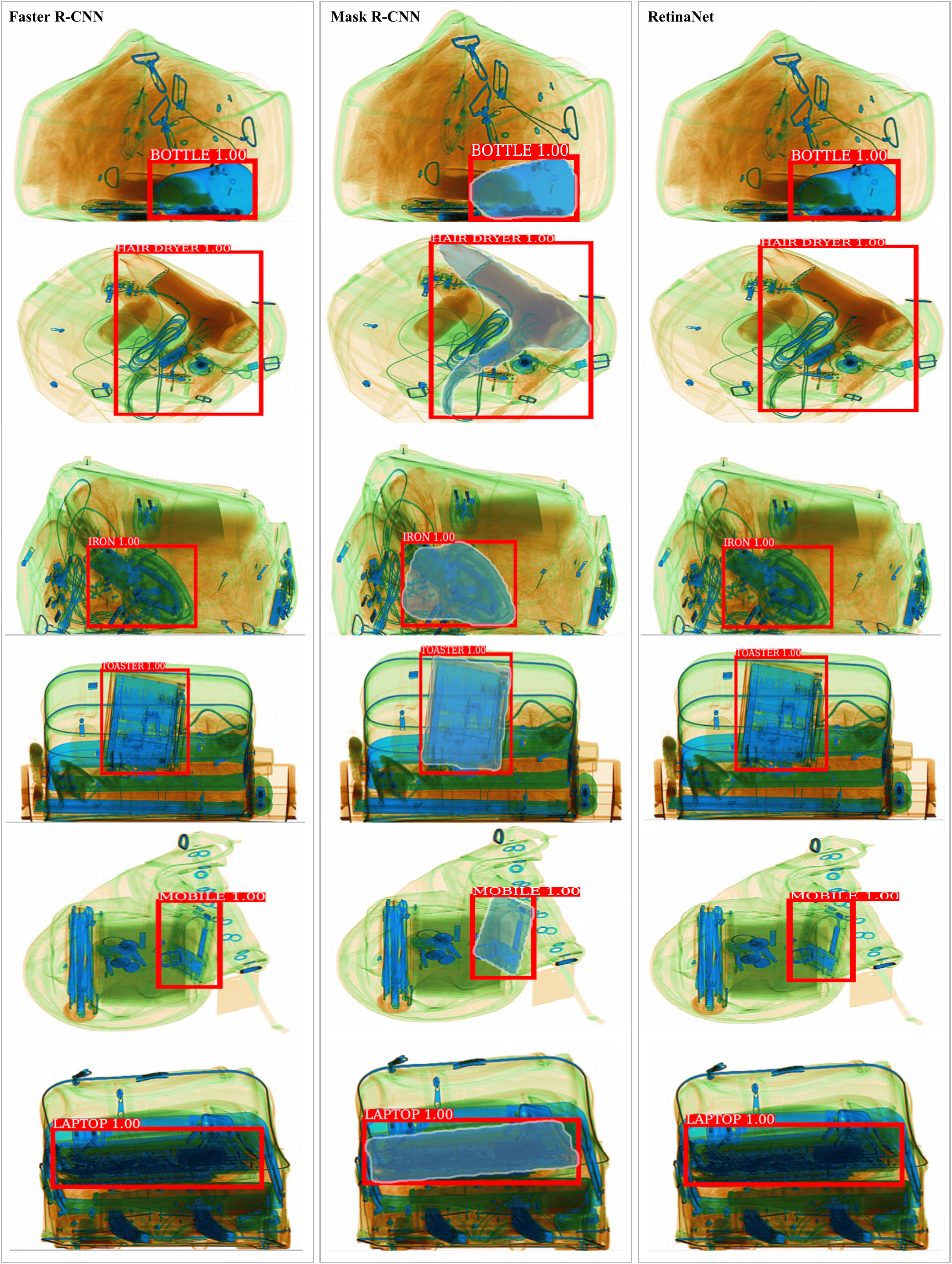}
    \caption{Examples of detection and classification of anomalous objects in X-ray security imagery using Faster R-CNN, Mask R-CNN and RetinaNet.}
    \label{fig:example}
\end{figure*}

\section{Results and Discussion}
\label{Section:Experimental Results and Discussion}

Results are presented for each states of our dual CNN architecture:- (a) primary object detection (Table \ref{Table:detection AP and mAP}); and (b) secondary classification of each detected object via two-class, $\{anomaly, benign\}$, classification (Table \ref{Table:binary class})

%%%%%%%%%%%%%%%%%%%%%%%%%%%%%%%%%%%%%%%%%%%%%%%%%%%%%%%%%%%%%%%%

\subsection{Object Detection Results}
Table \ref{Table:detection AP and mAP} presents object detection results for six object types (classes) using Faster R-CNN, Mask R-CNN and RetinaNet applied to the X-ray security imagery dataset outlined in Section \ref{Subsection:Datasets}. The highlighted AP/mAP signifies the maximal results obtained for each object class. 

% However, there are some examples where certain CNN based detection fails to classify threats, i.e. Faster R-CNN and Mask RCNN can detect laptop with high confidence while failing to detect when using RetinaNet, as shown in Fig. \ref{Fig:model_detection_compare}. 
However, there are where certain CNN based detection fails to classify threats, i.e., RetinaNet fails to detect laptop (as illustrated in Fig. \ref{Fig:model_detection_compare}C), while Faster R-CNN and Mask R-CNN can detect the laptop with high confidence (Figs. \ref{Fig:model_detection_compare}A and \ref{Fig:model_detection_compare}B).
Overall, the Mask R-CNN architecture gives superior performance, with a ResNet$_{101}$ backbone network giving superior performance across all architectures. The Mask R-CNN with ResNet$_{101}$ yields the highest mAP of 97.9\%  which establishes a new benchmark for object instance segmentation within X-ray security imagery.

%%%%%%%%%%%%%%%%%%%%%%%%%%%%%%%%%%%%%%%%%%%%%%%%%%%%%%%%%%%%%%%%
\subsection{Anomaly Detection Results}

Table \ref{Table:binary class} presents a side-by-side comparison of our secondary anomaly detection strategy (Section \ref{Subsection:Classification Strategies}) operating on objects that have been pre-localized (segmented) via our earlier object detection approach (dual CNN, Table \ref{Table:binary class} -- upper) against a simplistic approach of processing the full X-ray image containing the object without any prior object localization  (full image, Table \ref{Table:binary class} -- lower). Our use of the proposed fine-grain classification approach is additionally presented for the former case, where it represents a fine-grained sub-object classification problem.

From Table \ref{Table:binary class} we can observe superior performance in terms of statistical accuracy (A), precision (P), recall (R) and true positive (TP) with the dual CNN architecture that offers pre-localization of the objects from the image.  This observation demonstrates that dual CNN model can satisfactorily leverage the mutual benefits of the two complementary networks. By having object localisation in the first stage, it effectively makes the feature representation more meaningful (i.e. {\it focused}) for the secondary $\{anomaly, benign\}$ classification task. 
 
However, despite this success, we additionally note a high number of false positives (FP) presented across the Table \ref{Table:binary class}. Interestingly, the lowest FP comes from a full image rather than a dual CNN processing pipeline (albeit with lower overall performance than others).

Qualitative examples of the detection and classification of the various objects are presented in Fig. \ref{fig:example}. Our approach benefits from the performance of some classes which usually easy to distinguish by size and shape (e.g. hairdryer) while performance is lessor on smaller items, as shown in Table \ref{Table:detection AP and mAP}.

Overall, these results illustrate the challenge of anomaly detection within X-ray security imagery at both an image or object level.

%%%%%%%%%%%%%%%%%%%%%%%%%%%%%%%%%%%%%%%%%%%%%%%%%%%%%%%%%%%%%%%%

    %%%%%%%%%%%%%%%%%%%%%%%%%%%%%%%%%%%%%%%%%%%%%%%%%%%%%%%%%%%%%%%%

\section{Conclusion}
\label{Section:Conclusion}

In this work, we evaluate the effectiveness of dual CNN architecture for anomaly detection in the multiple-class item, \{{\it bottle, hairdryer, iron, toaster, mobile, laptop}\} in cluttered X-ray security imagery. We focus on two sub-problems: firstly, to leverage recent advances object detection to provide object localisation of threat item and secondly, leveraging established CNN and fine-grained classification to determine anomaly or benign objects. Experimentation demonstrates that fine-tuning of Mask R-CNN with ResNet$_{101}$ for X-ray imagery yields 97.9\% mAP for the first stage of object detection. However, while experimental results on secondary anomaly detection via a two-class classification problem, \{{\it anomaly, benign}\} show the benefits of a dual CNN architecture (TP: 76.86\% Accuracy: 66\%) false positive detection remains a significant issue (FP $\geqslant$10\%). Overall this illustrates the challenges of considering anomaly detection as an object-wise classification problem, even with recent advances in object detection within X-ray security imagery \cite{Akcay2018Xray}, remain significant when considering existing non-specialised CNN architectures for this task.
\\
\\
\noindent{\bf Acknowledgements}: Funding support - UK Department of Transport, Future Aviation Security Solutions (FASS) programme, (2018/2019)

%%%%%%%%%%%%%%%%%%%%%%%%%%%%%%%%%%%%%%%%%%%%%%%%%%%%%%%%%%%%%%%%
    %%%%%%%%%%%%%%%%%%%%%%%%%%%%%%%%%%%%%%%%%%%%%%%%%%%%%%%%%%%%%%%%

\bibliographystyle{IEEEtran}
\bibliography{ref/reference}

\begin{thebibliography}{10}
\providecommand{\url}[1]{#1}
\csname url@samestyle\endcsname
\providecommand{\newblock}{\relax}
\providecommand{\bibinfo}[2]{#2}
\providecommand{\BIBentrySTDinterwordspacing}{\spaceskip=0pt\relax}
\providecommand{\BIBentryALTinterwordstretchfactor}{4}
\providecommand{\BIBentryALTinterwordspacing}{\spaceskip=\fontdimen2\font plus
\BIBentryALTinterwordstretchfactor\fontdimen3\font minus
  \fontdimen4\font\relax}
\providecommand{\BIBforeignlanguage}[2]{{%
\expandafter\ifx\csname l@#1\endcsname\relax
\typeout{** WARNING: IEEEtran.bst: No hyphenation pattern has been}%
\typeout{** loaded for the language `#1'. Using the pattern for}%
\typeout{** the default language instead.}%
\else
\language=\csname l@#1\endcsname
\fi
#2}}
\providecommand{\BIBdecl}{\relax}
\BIBdecl

\bibitem{Rogers2016ReviewXray}
T.~W. Rogers, N.~Jaccard, E.~J. Morton, and L.~D. Griffin, ``Automated x-ray
  image analysis for cargo security: Critical review and future promise,''
  \emph{Journal of X-ray Science and Technology}, vol.~25, no.~1, pp. 33--56,
  2017.

\bibitem{Akcay2016Xray}
S.~Ak\c{c}ay, M.~E. Kundegorski, M.~Devereux, and T.~P. Breckon, ``Transfer
  learning using convolutional neural networks for object classification within
  x-ray baggage security imagery,'' in \emph{IEEE International Conference on
  Image Processing}, Sept 2016, pp. 1057--1061.

\bibitem{Akcay2017Xray}
S.~Ak\c{c}ay and T.~P. Breckon, ``An evaluation of region based object
  detection strategies within x-ray baggage security imagery,'' in \emph{IEEE
  International Conference on Image Processing}, Sept 2017, pp. 1337--1341.

\bibitem{Akcay2018Xray}
S.~Ak\c{c}ay, M.~E. Kundegorski, C.~G. Willcocks, and T.~P. Breckon, ``Using
  deep convolutional neural network architectures for object classification and
  detection within x-ray baggage security imagery,'' \emph{IEEE Transactions on
  Information Forensics and Security}, vol.~13, no.~9, pp. 2203--2215, Sept
  2018.

\bibitem{government2015digital}
``Hand luggage restrictions at uk airports,''
  \url{https://www.gov.uk/hand-luggage-restrictions}, accessed: 2019-03-13.

\bibitem{singh2003explosives}
S.~Singh and M.~Singh, ``Explosives detection systems (eds) for aviation
  security,'' \emph{Signal Processing}, vol.~83, no.~1, pp. 31--55, 2003.

\bibitem{wells2012review}
K.~Wells and D.~Bradley, ``A review of x-ray explosives detection techniques
  for checked baggage,'' \emph{Applied Radiation and Isotopes}, vol.~70, no.~8,
  pp. 1729--1746, 2012.

\bibitem{korupski2018evaluation}
J.~Skorupski and P.~Uchro{\'n}ski, ``Evaluation of the effectiveness of an
  airport passenger and baggage security screening system,'' \emph{Journal of
  Air Transport Management}, vol.~66, pp. 53--64, 2018.

\bibitem{Patcha2007:AnomalyOverview}
A.~Patcha and J.~Park, ``An overview of anomaly detection techniques: Existing
  solutions and latest technological trends,'' \emph{Computer Networks},
  vol.~51, no.~12, pp. 3448 -- 3470, 2007.

\bibitem{Schlegl2017:AnomalyGAN}
T.~Schlegl, P.~Seeb{\"o}ck, S.~M. Waldstein, U.~Schmidt-Erfurth, and G.~Langs,
  ``Unsupervised anomaly detection with generative adversarial networks to
  guide marker discovery,'' in \emph{Information Processing in Medical
  Imaging}.\hskip 1em plus 0.5em minus 0.4em\relax Cham: Springer International
  Publishing, 2017, pp. 146--157.

\bibitem{Kiran2018:AnomalyVideo}
B.~R. Kiran, D.~M. Thomas, and R.~Parakkal, ``An overview of deep learning
  based methods for unsupervised and semi-supervised anomaly detection in
  videos,'' \emph{Journal of Imaging}, vol.~4, no.~2, 2018.

\bibitem{greenemeier_2010}
\BIBentryALTinterwordspacing
L.~Greenemeier, ``Exposing the weakest link: As airline passenger security
  tightens, bombers target cargo holds,'' \emph{Scientific American}, Nov 2010.
  [Online]. Available:
  \url{https://www.scientificamerican.com/article/aircraft-cargo-bomb-security/}
\BIBentrySTDinterwordspacing

\bibitem{brown_2018}
\BIBentryALTinterwordspacing
M.~Brown, ``Brothers plead not guilty to meat grinder bomb plot,'' \emph{ABC
  News}, May 2018. [Online]. Available:
  \url{https://www.abc.net.au/news/2018-05-04/brothers-accused-of-plotting-to-blow-up-plane-plead-not-guilty/9726952}
\BIBentrySTDinterwordspacing

\bibitem{he2017maskrcnn}
K.~{He}, G.~{Gkioxari}, P.~{Dollár}, and R.~{Girshick}, ``Mask r-cnn,'' in
  \emph{2017 IEEE International Conference on Computer Vision}, Oct 2017, pp.
  2980--2988.

\bibitem{bastan2011xray}
M.~Ba{\c{s}}tan, M.~R. Yousefi, and T.~M. Breuel, ``Visual words on baggage
  x-ray images,'' in \emph{Computer Analysis of Images and Patterns}.\hskip 1em
  plus 0.5em minus 0.4em\relax Springer Berlin Heidelberg, 2011, pp. 360--368.

\bibitem{turcsany2013xray}
D.~Turcsany, A.~Mouton, and T.~P. Breckon, ``Improving feature-based object
  recognition for x-ray baggage security screening using primed visualwords,''
  in \emph{IEEE International Conference on Industrial Technology}, Feb 2013,
  pp. 1140--1145.

\bibitem{Bastan2013ObjectRI}
M.~Ba{\c{s}}tan, W.~Byeon, and T.~M. Breuel, ``Object recognition in multi-view
  dual energy x-ray images.'' in \emph{British Machine Vision Conference},
  vol.~1, no.~2, 2013, p.~11.

\bibitem{kundegorski2016xray}
M.~Kundegorski, S.~Ak\c{c}ay, M.~Devereux, A.~Mouton, and T.~Breckon, ``On
  using feature descriptors as visual words for object detection within x-ray
  baggage security screening.'' in \emph{International Conference on Imaging
  for Crime Detection and Prevention}.\hskip 1em plus 0.5em minus 0.4em\relax
  IET, January 2016, p. 12(6.).

\bibitem{Krizhevsky2017ImageNet}
A.~Krizhevsky, I.~Sutskever, and G.~E. Hinton, ``Imagenet classification with
  deep convolutional neural networks,'' \emph{Commun. ACM}, vol.~60, no.~6, pp.
  84--90, May 2017.

\bibitem{he2016residual}
K.~He, X.~Zhang, S.~Ren, and J.~Sun, ``Deep residual learning for image
  recognition,'' in \emph{IEEE Conference on Computer Vision and Pattern
  Recognition}, June 2016, pp. 770--778.

\bibitem{szegedy2016inception}
C.~Szegedy, S.~Ioffe, V.~Vanhoucke, and A.~A. Alemi, ``Inception-v4,
  inception-resnet and the impact of residual connections on learning,'' in
  \emph{AAAI Conference on Artificial Intelligence}, 2017.

\bibitem{howard2017mobilenets}
A.~G. Howard, M.~Zhu, B.~Chen, D.~Kalenichenko, W.~Wang, T.~Weyand,
  M.~Andreetto, and H.~Adam, ``Mobilenets: Efficient convolutional neural
  networks for mobile vision applications,'' \emph{CoRR}, vol. abs/1704.04861,
  2017.

\bibitem{Jaccard2016a}
N.~Jaccard, T.~Rogers, E.~Morton, and L.~Griffin, ``{Automated detection of
  smuggled high-risk security threats using Deep Learning},'' in \emph{7th
  International Conference on Imaging for Crime Detection and
  Prevention}.\hskip 1em plus 0.5em minus 0.4em\relax Institution of
  Engineering and Technology, 3 2016, pp. 11 (4 .)--11 (4 .).

\bibitem{Mery2017Xray}
D.~Mery, E.~Svec, M.~Arias, V.~Riffo, J.~M. Saavedra, and S.~Banerjee, ``Modern
  computer vision techniques for x-ray testing in baggage inspection,''
  \emph{IEEE Transactions on Systems, Man, and Cybernetics: Systems}, vol.~47,
  no.~4, pp. 682--692, April 2017.

\bibitem{Sterchi2017}
Y.~Sterchi, N.~H{\"{a}}ttenschwiler, S.~Michel, and A.~Schwaninger,
  ``{Relevance of visual inspection strategy and knowledge about everyday
  objects for X-ray baggage screening},'' in \emph{International Carnahan
  Conference on Security Technology}, 2017, pp. 1--6.

\bibitem{griffin2018unexpected}
L.~D. Griffin, M.~Caldwell, J.~T. Andrews, and H.~Bohler, ``“unexpected item
  in the bagging area”: Anomaly detection in x-ray security images,''
  \emph{IEEE Transactions on Information Forensics and Security}, vol.~14,
  no.~6, pp. 1539--1553, 2019.

\bibitem{zheng2013xray}
Y.~Zheng and A.~Elmaghraby, ``A vehicle threat detection system using
  correlation analysis and synthesized x-ray images,'' in \emph{Detection and
  Sensing of Mines, Explosive Objects, and Obscured Targets XVIII}, vol.
  8709.\hskip 1em plus 0.5em minus 0.4em\relax International Society for Optics
  and Photonics, 2013, p. 87090V.

\bibitem{Andrews2016:AnomalyAutoEncoder}
J.~Andrews, E.~J. Morton, and L.~D. Griffin, ``{Detecting Anomalous Data Using
  Auto-Encoders},'' \emph{International Journal of Machine Learning and
  Computing}, vol.~6, no.~1, pp. 21--26, 2016.

\bibitem{andrews2017xray}
J.~T. Andrews, N.~Jaccard, T.~W. Rogers, and L.~D. Griffin,
  ``Representation-learning for anomaly detection in complex x-ray cargo
  imagery,'' in \emph{Anomaly Detection and Imaging with X-Rays (ADIX) II},
  vol. 10187.\hskip 1em plus 0.5em minus 0.4em\relax International Society for
  Optics and Photonics, 2017, p. 101870E.

\bibitem{Akcay2018GANomaly}
S.~Akcay, A.~Atapour-Abarghouei, and T.~P. Breckon, ``Ganomaly :
  semi-supervised anomaly detection via adversarial training.'' in \emph{14th
  Asian Conference on Computer Vision}.\hskip 1em plus 0.5em minus 0.4em\relax
  Springer, December 2018.

\bibitem{Akcay2019}
S.~Ak{\c{c}}ay, A.~Atapour-Abarghouei, and T.~P. Breckon, ``Skip-ganomaly: Skip
  connected and adversarially trained encoder-decoder anomaly detection,''
  \emph{arXiv preprint arXiv:1901.08954}, 2019.

\bibitem{ren2015fasterrcnn}
S.~Ren, K.~He, R.~Girshick, and J.~Sun, ``Faster r-cnn: Towards real-time
  object detection with region proposal networks,'' in \emph{Advances in neural
  information processing systems}, 2015, pp. 91--99.

\bibitem{lin2017focalloss}
T.-Y. Lin, P.~Goyal, R.~Girshick, K.~He, and P.~Doll{\'a}r, ``Focal loss for
  dense object detection,'' in \emph{Proceedings of the IEEE international
  conference on computer vision}, 2017, pp. 2980--2988.

\bibitem{redmon2016you}
J.~Redmon, S.~Divvala, R.~Girshick, and A.~Farhadi, ``You only look once:
  Unified, real-time object detection,'' in \emph{Proc. Conf. on Computer
  Vision and Pattern Recognition}.\hskip 1em plus 0.5em minus 0.4em\relax IEEE,
  2016, pp. 779--788.

\bibitem{liu2015ssd}
W.~Liu, D.~Anguelov, D.~Erhan, C.~Szegedy, S.~Reed, C.-Y. Fu, and A.~C. Berg,
  ``Ssd: Single shot multibox detector,'' in \emph{European Conference on
  Computer Vision}.\hskip 1em plus 0.5em minus 0.4em\relax Springer, 2016, pp.
  21--37.

\bibitem{Iandola2016SqueezeNet}
F.~N. Iandola, S.~Han, M.~W. Moskewicz, K.~Ashraf, W.~J. Dally, and K.~Keutzer,
  ``Squeezenet: Alexnet-level accuracy with 50x fewer parameters and
  {\textless} 0.5 mb model size,'' \emph{arXiv preprint arXiv:1602.07360},
  2016.

\bibitem{simonyan2014verydeep}
K.~Simonyan and A.~Zisserman, ``Very deep convolutional networks for
  large-scale image recognition,'' \emph{arXiv preprint arXiv:1409.1556}, 2014.

\bibitem{Russakovsky2015ImageNet}
O.~Russakovsky, J.~Deng, H.~Su, J.~Krause, S.~Satheesh, S.~Ma, Z.~Huang,
  A.~Karpathy, A.~Khosla, M.~Bernstein, A.~C. Berg, and L.~Fei-Fei, ``Imagenet
  large scale visual recognition challenge,'' \emph{International Journal of
  Computer Vision}, vol. 115, no.~3, pp. 211--252, Dec 2015.

\bibitem{Wang2018LearningAD}
Y.~Wang, V.~I. Morariu, and L.~S. Davis, ``Learning a discriminative filter
  bank within a cnn for fine-grained recognition,'' in \emph{Proceedings of the
  IEEE Conference on Computer Vision and Pattern Recognition}, 2018, pp.
  4148--4157.

\bibitem{VanHorn2015iIrds}
G.~V. Horn, S.~Branson, R.~Farrell, S.~Haber, J.~Barry, P.~Ipeirotis,
  P.~Perona, and S.~Belongie, ``Building a bird recognition app and large scale
  dataset with citizen scientists: The fine print in fine-grained dataset
  collection,'' in \emph{IEEE Conference on Computer Vision and Pattern
  Recognition}, June 2015, pp. 595--604.

\bibitem{Wah2011Birds}
C.~Wah, S.~Branson, P.~Welinder, P.~Perona, and S.~Belongie, ``The caltech-ucsd
  birds-200-2011 dataset,'' Tech. Rep., 2011.

\bibitem{Khosla2011finegrained}
A.~Khosla, N.~Jayadevaprakash, B.~Yao, and L.~Fei-Fei, ``Novel dataset for
  fine-grained image categorization,'' in \emph{First Workshop on Fine-Grained
  Visual Categorization, IEEE Conference on Computer Vision and Pattern
  Recognition}, Colorado Springs, CO, June 2011.

\bibitem{Wegner2016Catalogue}
J.~D. Wegner, S.~Branson, D.~Hall, K.~Schindler, and P.~Perona, ``Cataloging
  public objects using aerial and street-level images-urban trees,'' in
  \emph{Proceedings of the IEEE Conference on Computer Vision and Pattern
  Recognition}, 2016, pp. 6014--6023.

\bibitem{Mouton2015ARO}
A.~Mouton and T.~P. Breckon, ``A review of automated image understanding within
  3d baggage computed tomography security screening.'' \emph{Journal of X-ray
  science and technology}, vol. 23 5, pp. 531--55, 2015.

\bibitem{gilardoni_scanner}
``Gilardoni- x-ray and ultrasounds,'' \url{https://www.gilardoni.it/en/},
  accessed: 2019-03-13.

\bibitem{lin2014coco}
T.-Y. Lin, M.~Maire, S.~Belongie, J.~Hays, P.~Perona, D.~Ramanan,
  P.~Doll{\'a}r, and C.~L. Zitnick, ``Microsoft coco: Common objects in
  context,'' in \emph{European Conference on Computer Vision}.\hskip 1em plus
  0.5em minus 0.4em\relax Springer, 2014, pp. 740--755.

\bibitem{Girshick2018Detectron}
R.~Girshick, I.~Radosavovic, G.~Gkioxari, P.~Doll\'{a}r, and K.~He,
  ``Detectron,'' \url{https://github.com/facebookresearch/detectron}, 2018.

\bibitem{deng2009imagenet}
J.~Deng, W.~Dong, R.~Socher, L.-J. Li, K.~Li, and L.~Fei-Fei, ``Imagenet: A
  large-scale hierarchical image database,'' in \emph{IEEE Conference on
  Computer Vision and Pattern Recognition}.\hskip 1em plus 0.5em minus
  0.4em\relax IEEE, 2009, pp. 248--255.

\end{thebibliography}

%%%%%%%%%%%%%%%%%%%%%%%%%%%%%%%%%%%%%%%%%%%%%%%%%%%%%%%%%%%%%%%%
    
    % \listoftodos
\end{document}